\documentclass{article} 

\usepackage{microtype}
\usepackage{graphicx}
\usepackage{subfigure}
\usepackage{booktabs} 
\usepackage{float}

\usepackage{hyperref}

\usepackage[accepted]{icml2021}

\usepackage[shortlabels]{enumitem}

\usepackage{times}
\usepackage{soul}
\usepackage{url}
\usepackage{caption}
\usepackage{amsmath}
\usepackage{amsthm}
\usepackage{amssymb}
\usepackage{amsfonts}
\usepackage{booktabs}
\usepackage{algorithm}
\usepackage{algorithmic}
\newtheorem{theorem}{Theorem}
\newtheorem{lemma}{Lemma}

\usepackage{helvet}  
\usepackage{courier}  
\usepackage{subfigure}
\usepackage{amsfonts}       
\usepackage{nicefrac}       
\usepackage{amsmath}
\usepackage{stfloats}

\newcommand{\beqn}{\begin{eqnarray}}
\newcommand{\eeqn}{\end{eqnarray}}
\newcommand{\beqnx}{\begin{eqnarray*}}
	\newcommand{\eeqnx}{\end{eqnarray*}}

\newcommand{\tabref}[1]{Table~\ref{#1}}

\newcommand{\figref}[1]{Fig.~\ref{#1}}
\newcommand{\lemref}[1]{Lemma~\ref{#1}}
\newcommand{\thmref}[1]{Theorem~\ref{#1}}

\icmltitlerunning{Stochastic Boundary Ordinary Differential Equation}

\begin{document}

\twocolumn[
\icmltitle{STRODE: Stochastic Boundary Ordinary Differential Equation  }




\begin{icmlauthorlist}
	\icmlauthor{Hengguan Huang}{1}
	\icmlauthor{Hongfu Liu}{1}
	\icmlauthor{Hao Wang}{2}
	\icmlauthor{Chang Xiao}{1}
	\icmlauthor{Ye Wang}{1}
\end{icmlauthorlist}

\icmlaffiliation{1}{National University of Singapore}
\icmlaffiliation{2}{Rutgers University}

\icmlcorrespondingauthor{Hengguan Huang}{hengguan@comp.nus.edu.sg}
\icmlcorrespondingauthor{Ye Wang}{wangye@comp.nus.edu.sg}

\icmlkeywords{Neural Ordinary differential equation,Boundary value problem, Point process, Postdiction }

\vskip 0.3in
]



\printAffiliationsAndNotice{} 

\begin{abstract}
Perception of time from sequentially acquired sensory inputs is rooted in everyday behaviors of individual organisms. Yet, most algorithms for time-series modeling fail to learn dynamics of random event timings directly from visual or audio inputs, requiring timing annotations during training that are usually unavailable for  real-world applications. For instance, neuroscience perspectives on postdiction imply that there exist variable temporal ranges within which the incoming sensory inputs can affect the earlier perception, but such temporal ranges are mostly unannotated for real applications such as automatic speech recognition (ASR).
In this paper, we present a probabilistic ordinary differential equation (ODE), called STochastic boundaRy ODE (STRODE\footnote{Our code is available online: \url{https://github.com/Waffle-Liu/STRODE} }), that learns both the timings and the dynamics of time series data without requiring any timing annotations during training. STRODE allows the usage of differential equations to sample from the posterior point processes, efficiently and analytically. We further provide theoretical guarantees on the learning of STRODE. Our empirical results show that our approach successfully infers event timings of time series data.
Our method achieves competitive or superior performances compared to existing
state-of-the-art methods for both synthetic and real-world datasets.

\end{abstract}

\section{Introduction}

Perception of time from sequentially acquired sensory inputs is rooted in the everyday behavior of the individual organism. 
Numerous neuroscience studies have provided evidence for close connections between time perception and sensory input from multiple sensory modalities such as audition and vision \cite{bolognini2012hearing,murai2018optimal,ulrich1998effects}. Decoupling the perception of time and content from sensory inputs is crucial for real-time perception in fast-changing environments \cite{plos2014correction,toso2020sensory}. For example, to perform interceptive actions such as hitting a moving object, precise timing is required due to significant response latency of neurons; to identify ambiguous phoneme from the acoustic signal in real-time, again, the timing is required such that the human brain is able to strike a balance between postdiction (to ensure sufficient information is integrated from subsequent context) and prediction (to ensure decision outcomes are not delayed relative to the external world) \cite{gwilliams2018spoken}.

Despite the enormous practical successes of machine learning (in particular, deep learning),  most of the algorithms for time-series modeling fail to learn dynamics of random event timings directly from visual or audio inputs, and still require their training data to have timing annotations, such as Latent ODE \cite{rubanova2019latent}.
However, for some real-world applications in which timing annotations are not available, it is necessary to introduce a ``time perception'' mechanism to handle such uncertainty.  For instance, neuroscience perspectives on postdiction imply that there exist variable temporal ranges within which the incoming sensory inputs can affect the earlier perception, but such temporal ranges are mostly unannotated for real applications such as automatic speech recognition (ASR) \cite{yu2018recent}. 


Can gaps between natural and artificial intelligence be bridged further through introducing the ``time perception'' mechanism? In this paper, we generalize neural ODE in handling a special type of boundary value problem with random boundary times which are described by a temporal point process (TPP). We present a probabilistic ordinary differential equation (ODE) called stochastic boundary ODE (STRODE) that infers both the timings and the dynamics of the time series data without requiring any timing annotations during training.

We adopt variational inference to optimize our model. The boundary time variables involved in STRODE are described as ODEs, consequently posing a major challenge for evaluating  evidence lower bound (ELBO)  with respect to sampling and inference of such distributions. To mitigate this challenge, we propose a method that allows for joint inference and differentiable sampling of such distributions through solving ODEs. Furthermore, the KL term between two differential equations in the ELBO is computationally intractable. 
We further provide an analytical upper bound for the KL term such that we have a closed-form solution for the ELBO.

Our empirical results show that our approach successfully infers event timings of the time series data.  The experiments over Rotating MNIST Thumbnail, a synthetic video thumbnail dataset, show that our model is capable of
inferring event timings of the irregularly sampled high-dimensional data without using timing annotations during training, whereby 
learning of the complex dynamics of irregularly sampled data is achieved. We further apply our model for postdictive modeling using CHiME-5 speech data. We demonstrate that our new model  outperforms baseline neural ODEs.   

\section{Related Work}
\label{bg}

Ordinary differential equations (ODE) are powerful mathematical tools to describe continuous-time dynamics of an evolving
system such as chemical transformation in chemistry \cite{verwer1995explicit}, and laws of motion in physics and disease spreading \cite{komarova2010ode} in biology. They have been recently applied to model dynamics of hidden representations of neural networks \cite{chen2018neural}. Such neural ODE models have increasingly gained attention in the machine learning community. For example,   \cite{grathwohl2018ffjord} improves upon neural ODE in terms of efficiency by introducing an unbiased stochastic estimator of the likelihood. Since neural ODE assumes the data to be evenly distributed, ODE-recurrent neural network (ODE-RNN) \cite{rubanova2019latent} further extends this approach to handle irregularly sampled data by describing state transitions in recurrent neural network (RNN) as an initial value problem (IVP).  Along a different line of research, jump stochastic differential equations (JSDE) \cite{jia2019neural} incorporate a temporal point process into neural ODE to model marked point processes. Though these variants of neural ODE are capable of learning dynamics of time series data with or without discontinuities, they require timing annotations for each data sample. Therefore, they are not applicable to
our tasks where timing annotations are not available during training. 

The closest work to ours is \cite{chen2020learning}, in which neural event functions are introduced in neural ODE solvers to enable the learning of termination criteria. However, the design of such event function requires prior knowledge of the dynamical system. In contrast, our model adopts a regenerative point process as prior, whose parameters are learned from data itself. Furthermore, \cite{chen2020learning}  focuses on solving IVPs, while our approach generalizes neural ODE in handling a special type of boundary value problem with random boundary conditions.  

\section{Background }

\subsection{Temporal Point process}
A temporal point process (TPP) is a stochastic process that describes the temporal dependence among events. It provides an effective solution to solve the next-event-prediction problem. A TPP  can be equivalently represented as multiple sets of random variables such as arrival times  and  waiting times. Let $\{ t_i\}_{i=0}^{N}$ be a sequence of arrival times and let $\{ \Delta{ t_i}\}_{i=1}^{N} $ be a sequence of waiting times sampled from a TPP  $T(\{p_i(t) \}_{i=0}^{N}  )$, we have:
\begin{align}
   \{ t_i \}_{i=0}^{N}   \sim T(\{p_i(t) \}_{i=0}^{N} )  \\
    t_{i}  \sim p_i(t) \\
    \Delta{t_{i}} = t_{i}-t_{i-1} 
\end{align}
where $p_i(t)$ denotes the probability density function of the arrival time $t_i$.  All these random variables involved in TPP are characterized by a conditional intensity function $\lambda(t)$, which is defined to be the rate at which events are expected to occur at time $t$ given histories. For example, the probability density $p_i(t)$ is written as:
\begin{equation}
p_{i}(t)=  \lambda(t) e^{\int_{t_{i-1}}^{t}{-\lambda(t)dt}}
\end{equation}

Given the last arrival time $t_{i-1}$, the expected next arrival time can be generated by:
\begin{equation} \label{sample}
    \Bar{t}_i = \int_{t_{i-1}}^{+\infty} t p_{i}(t) dt
\end{equation}

Let the first arrival time be placed at $t_0=0$. If the process follows a Poisson process or more generally a regenerative process, it probabilistically restarts at any arrival time, i.e., for any $i$, we have
\begin{equation}
p_{i}(t)= p_{0}(t) = \lambda(t) e^{\int_{0}^{t}{-\lambda(t)dt}}
\end{equation}

Parameter learning for TPP can be conducted by maximum likelihood estimation (MLE). Note that in this work, the arrival times of data samples are \emph{not given} during training. Therefore it is impossible to directly cast it as a supervised learning problem that could be solved via MLE. 



\subsection{From Initial Value Problem to Boundary Value Problem}
Ordinary differential equations (ODE),  initially exploited to describe the phenomena in  physical domain, have been applied to model dynamics of hidden representations of neural networks \cite{chen2018neural,rubanova2019latent}. For simplicity, consider an autoregressive task which predicts future value based on histories. Suppose we are given a sequence of training samples  $\textbf{X} =\{ x_i\}_{i=0}^{N} $ at times  $\{ t_i\}_{i=0}^{N} $ and let $t_{0}=0$.  For $i<N$ and $t \in [ t_{i},t_{i+1}] $, the hidden state $h(t)$ satisfies an initial value problem (IVP):
\begin{equation}
    {h}'(t) = f_{\theta_1}(h(t),t ),\    h(t_{i})=x_{i}
\end{equation}
where $f_{\theta_1}$  is a neural network. The general solution of $h(t_{i+1})$:
\begin{equation} 
    h(t_{i+1})  = h(t_{i})+ \int_{t_{i}}^{t_{i+1}} f_{\theta_1}(h(t),t ) dt 
\end{equation}
where $ h(t_{i+1})$ can be approximated using the ODE numerical solvers such as the Euler method and the Runge-Kutta methods  \cite{chen2018neural}, i.e.:
\begin{equation} \label{IVP_so}
\tilde{h}(t_{i+1}) = \mathrm{ODESolve}(f_{\theta_1},h(t_{i}), t_{i},t_{i+1} )
\end{equation}
Combining above $N$ IVPs at all time steps, for $t \in [ t_{0},t_{N}] $,  we have a boundary value problem (BVP) as follows:
\begin{equation}
     {h}'(t) = f_{\theta_1}(h(t),t )
\end{equation}
whose boundary conditions are:
\begin{align}
   \{ h(t_{0})=x_{0},h(t_{1})=x_{1} ,. . ., h(t_{N})=x_{N} \} 
\end{align}
To solve the above neural ODE that satisfies boundary conditions, we firstly follow Eq.\eqref{IVP_so} to construct a trial form of the  solution of $h(t)$ at each boundary time except $t_0$. Then the model can be optimized by minimizing the mean squared error (MSE):
\begin{equation}
\frac{1} {N} \sum_{i=1}^{N}  (\tilde{h}(t_{i})  -x_i)^{2}
\end{equation}
It can also be conducted by MLE. More specifically,
by assuming that each boundary value $x_i$  follows a Gaussian distribution $ p(x_{i}  )=\mathcal{N}(\mu_i,\sigma_{i}^{2})$  whose mean $\mu_i$ and variance $\sigma_{i}^{2}$ are the output of a  neural network,  we have:
\begin{equation}
     x_{i} \sim \mathcal{N}(\mu_i,\sigma_{i}^{2}), \ where \ [\mu_i,\sigma_{i}^{2} ] = f_{\theta_2}(\tilde{h}(t_i) )  
\end{equation}
It follows that the log-likelihood of all boundary values is: 
\begin{equation}
  \log P( \textbf{X})=\sum_{i=1}^{N} \log p(x_{i}  )
\end{equation}
Both IVPs and BVPs assume timings of data samples are available during training and thus restricting its capability in handling real-world problems such as postdiction-based ASR, where postdictive temporal ranges are unannotated.   

\section{Stochastic Boundary Ordinary Differential Equation (STRODE)}

\subsection{Stochastic Boundary Value Problem }  
Stochastic boundary value problem (SBVP) is a special type of boundary value problem we propose in which the unobservable boundary conditions/times are described by a stochastic process, e.g. temporal point process.  Let us consider a more challenging autoregressive task: we are only given a sequence of training samples $\textbf{X} =\{ x_i\}_{i=0}^{N} $, in which the first data sample $x_0$ occurs at $t_{0}=0$, our goal is not just one step ahead prediction, but to infer both boundary times $\{ t_i\}_{i=1}^{N} $ and latent dynamics of the time-series data. 

Let $\textsl{T}(\{p_i(t|\mathbf{x}_{i}) \}_{i=1}^{N})$ be a temporal point process 
whose realization consists of a sequence of boundary times $\{ \tilde{t}_i\}_{i=1}^{N}$, each follows a density functions $p_i(t|\mathbf{x}_{i})$ conditioned on the data sample $x_i$. For $t \in [ 0,\tilde{t}_{N}] $, the hidden states $h(t)$ satisfies an SBVP:
\begin{equation}
     {h}'(t) = f_{\theta_o}(h(t),t )
\end{equation}
whose boundary conditions are:
\begin{align}
 &  \{ \tilde{t}_i\}_{i=1}^{N} \sim \textsl{T}(\{p_i(t|\mathbf{x}_{i}) \}_{i=1}^{N}) \\
 &  \{ h(0) =x_{0},h(\tilde{t}_1)=x_{1} , . . ., h(\tilde{t}_{N})=x_{N} \} 
\end{align}
where $f_{\theta_o}$ is a neural network.

\subsection{Learning}

The neural ODE that involves in SBVP is dubbed as stochastic boundary ordinary differential equation (STRODE).  Learning the parameters of neural networks and TPP of the STRODE is, equivalently, solving the SBVP. It is challenging due to the difficulty in estimating parameters of $\textsl{T}(\{p_i(t|x_{i}) \}_{i=1}^{N}) $ without having any observable boundary times. In this work, we adopt variational inference to optimize our model. Let $q_i(t|x_{i})$ be the approximate posterior of the boundary time conditioned on the data sample $x_{i}$ and let $p_i(t)$ be the corresponding prior. The  evidence lower bound (ELBO) can be written as:
\begin{equation} \label{elbo}
\begin{split}
 \log P(\mathbf{X}) \geqslant  \sum_{i=1}^{N} \{  \mathbb{E}_{ \tilde{t}_i \sim q_i(t|x_{i}) }   \log p( x_i|\tilde{t}_i)  \\ -\mathrm{KL} (q_i(t|x_{i}) || p_i(t))  \}
 \end{split}
\end{equation}
Most of the existing TPPs make strong assumptions about the generative processes of the event data and use a fixed parametric form of the intensity function, restricting the expressive power of the respective processes.  
Instead, our model adopts a general form of TPP in which the arrival time variable could be an arbitrary distribution, consequently posing a major challenge for evaluating ELBO with respect to sampling and inference of such distributions. In the following subsections, we will show how to obtain a closed-form solution of ELBO.



\subsection{ODE-based  Sampling and Inference of TPP} 

Sampling from a TPP  is usually performed via thinning algorithm \cite{ogata1981lewis}. However, such an operation is not differentiable as there is no guarantee that the samples drawn from the algorithm may converge to the exact gradient estimate of the ELBO. To tackle this problem, we model TPP as  ODEs such that sampling is achieved by solving such ODEs. More importantly, all operations are differentiable. 

Given the last boundary time $t_{i-1}$, the expected next boundary time is written as:
\begin{equation}
   \bar{t}_{i}  =  \int_{t_{i-1}}^{+ \infty}  t q_i(t|x_{i}) dt  
\end{equation}
Inspired by this, we derive an IVP whose solution allows us to obtain a sequence of boundary time samples $\{ \tilde{t}_i\}_{i=1}^{N} $.   To be more specific, suppose the initial boundary time is placed at $\tilde{t}_{0}=0$.  Given the last boundary time sample $\tilde{t}_{i-1}$, let $ t \in [ 0,\tilde{t}_{i-1}] $, the next boundary time $\Phi_i(t)$ satisfies an IVP:
\begin{equation} \label{bigphi} 
  {  \Phi_i}'(t) = - t q_i(t|x_{i}) ,\    {\Phi_i}(0)=  \int_{0}^{+ \infty}  t q_i(t|x_{i}) dt  
\end{equation}
Then the general solution of $\tilde{t}_{i}$ is written as:
\begin{equation} \label{IVP_sample}
    \tilde{t}_{i} = {\Phi_i}(t_{i-1}) = {\Phi_i}(0)+ \int_{0}^{\tilde{t}_{i-1}} - t q_i(t|x_{i}) dt \nonumber
\end{equation} 

The ODE numerical solver, e.g. Euler method fails to generate the approximate solution as ${\Phi_i}(0)$ is computationally intractable. We, therefore, adopt the neural network $f_{\theta_\Phi}$  to approximate the solution of $\tilde{t}_{i}$, such that:
\begin{equation} \label{sample22}
\tilde{t}_{i} = {\Phi_i}(\tilde{t}_{i-1}) = f_{\theta_\Phi} (\tilde{t}_{i-1}, x_i) \end{equation}
Notably, we  use $\tilde{t}_{i-1}$ as the bias of the last layer of $f_{\theta_\Phi}$ to guarantee that  $\tilde{t}_{i}$ is greater than $\tilde{t}_{i-1}$ (implementation details in the Supplement).

Then  ${\Phi_i}'(t)$ can be  obtained by differentiating ${\Phi_i}(t)$ with respect to $t$. It follows that the approximate posterior $q_i(t|x_{i})$ can be written as:
\begin{equation}
   q_i(t|x_{i}) =      \frac{{-\Phi_i}'(t)}{t}     
\end{equation}

\subsection{Prior Point Process} 
We impose a regenerative process prior on our STRODE. Inspired by \cite{chen2018neural,rubanova2019latent}, we model the cumulative conditional intensity function of the process as an initial value ODE. Let $t \in [ 0,+\infty) $, the cumulative conditional intensity function $\phi_i(t)$ satisfies an initial value problem (IVP):
\begin{equation}
     {\phi}'_i(t) =\lambda_i(t), \ {\phi}_i(0)=0 
\end{equation}
Since our regenerative process prior probabilistically restarts itself for any arrivals,  the general solution for conditional intensity function at $t$ can be written as 
\begin{equation}
    \phi_i(t) = \phi_0(t)=  \int_{0}^{t} \lambda_i(s)ds
\end{equation}
such that the prior $p_i(t)$ is reformulated as follows:
\begin{equation}
p_{i}(t)=  {\phi}'_i(t) e^{-\phi_i(t) }
\end{equation}
Solving the above IVP requires designing of a specific functional form of the intensity function $\lambda_i(t)$, which results in a TPP with limited expressive power. To alleviate this problem, instead of 
parameterizing $\lambda_i(t)$, we adopt a neural network $f_{\theta_\phi}$ to approximate $\phi_i(t)$.   To ensure that both $\lambda_i(t)$ and $\phi_i(t)$ are constrained to be positive, we adopt similar neural network architecture of \cite{omi2019fully}.


\subsection{Upper Bound of ODE-based Kullback–Leibler (KL) Divergence}\label{sec:upper}

With both $q_i(t|x_{i})$ and   $p_i(t)$ being defined as differential equations, the KL term of the ELBO defined in  Eq.\eqref{elbo} can be written as :
\begin{equation}
  \mathrm{KL} (q_i(t|x_{i}) || p_i(t)) = \int_{0}^{+\infty} \frac{{-\Phi_i}'(t)}{t} \log \frac{{-\Phi_i}'(t)}{t{\phi}'_i(t) e^{-\phi_i(t) }}  
\end{equation}
The above KL term is computationally intractable as the upper limit of integration approaches infinity, and it remains unknown whether such an improper integral converges. The following theorem provides an analytical upper bound for the KL term such that we have a closed-form solution for the ELBO.

\begin{theorem}\label{th1}
Suppose we are given two arbitrary distributions, $q(t)$ and $p(t)$ with $t \in [0,+\infty)$. Let $ m=-e^{-t}$. Let $\epsilon$ be a positive real constant and let $g:[-1,0)\to R $ be a continuous function. There exists a $G:[-1,0)\to [0, +\infty)$ that satisfies an initial value problem: 
\begin{align*}
 &{G}'(m) = g(m),\ \ G(-1)=0 
 \nonumber
 \\ &where \  g(m) = \frac{-q(- \log(-m)) }{m} \log  \frac{ q(- \log(-m))}{p(- \log(-m))}
\end{align*}
Such that as $\epsilon \rightarrow 0$,  we have:
\begin{equation}
\lim_{\epsilon \rightarrow 0}  (\mathrm{KL}(q(t) || p(t))  ) \leqslant \lim_{\epsilon \rightarrow 0}  (G(-\epsilon)
\\  +   \| G(-2\epsilon) - G(-\epsilon)  \| ) 
\nonumber
\end{equation}

\end{theorem}

In the proof of \thmref{th1} (Appendix 2), an IVP is introduced to assist  derivation of the upper bound of the KL divergence between two arbitrary distributions. Considering specific functional forms of the KL , let $m=-e^{-t}$, let $M= -\log(-m)$ and  let  $m \in [-1,0)$, such an IVP can be written as:
\begin{equation} \label{26}
 {G}'(m) = g(m),\ \ G(-1)=0 
\end{equation}
where $g(m)$ is written as:
\begin{equation}
    g(m) = \frac{{-\Phi_i}'(M)}{-mM} \log \frac{{\Phi_i}'(M)}{M{\phi}'_i(M) e^{-\phi_i(M) }} 
\end{equation}
Then \thmref{th1}  separates the general solution of the KL, $\lim_{l \rightarrow 0}G(l)$  into two:  $G(-\epsilon)$ and an improper integral: 
\begin{equation} \label{eqp}
   G(-\epsilon) + \lim_{l\rightarrow 0} \int_{-\epsilon}^{l } {g(m) dm}
\end{equation}
where $\epsilon$ in this work is set as the step size of the Euler method that we apply to calculate $G(-\epsilon)$. Obviously, the second term of Eq.~\eqref{eqp} can be written as an IVP. \thmref{th1} then introduces another IVP with delay $\epsilon$, consequently applying the following lemma to derive the upper bound of the second term.

\begin{lemma}\label{l1}
Let $\epsilon$ be a positive real constant. Let $U\subset R^{n}$ be an open set. Let $f_1,f_2:[a-2\epsilon,a)\to R^{n}$ be a continuously differentiable function and $\left \| {f_1}' \right \| \leqslant M$ where $M$ is a positive constant.   Let $y_1,y_2:[a-\epsilon,a)\to U$ satisfy the initial value problem: 
\begin{align*}
{y}'_{1} &= f_1(t), \ \ y_1(a-\epsilon)=x_1 \\
{y}'_{2} &= f_2(t) = f_1(t-\epsilon),\ \ y_2(a-\epsilon)=x_2 
\end{align*}
Suppose both $x_1$ and $x_2$ depend on $\epsilon$. As $\epsilon \rightarrow 0$, we have: 
\begin{equation*}
\lim_{\epsilon \rightarrow 0} 
\left (\lim_{t \rightarrow a} { \left \| {y}_{1}(t) - {y}_{2}(t) \right \|}\right ) \leqslant  \lim_{\epsilon \rightarrow 0}  \left \| x_1 - x_2 \right \|
\end{equation*}
\end{lemma}
The Lemma (proof given in Appendix 1) states that given two IVPs, one of which is imposed with a delay,  the difference between two terminal states is bounded by the difference between initial
states. In contrast, Gronwall’s Inequality (Theorem 2.1) \cite{howard1998gronwall} provides a bound involving an unbounded Lipschitz constant of the ODE, which is not applicable for our problem setting. Therefore, with \lemref{l1}, the second term in Eq.~\eqref{eqp} is upper bounded by:
\begin{equation} \label{ipi}
\lim_{l\rightarrow 0} \int_{-\epsilon}^{l } {g(m) dm} \leqslant  \left \| G(-2\epsilon) - G(-\epsilon) \right \|
\end{equation}
Combining Eq.~(\ref{26}-\ref{ipi}), the upper bound of the KL term of the ELBO is written as:
\begin{equation}
    \mathrm{KL} (q_i(t|x_{i}) || p_i(t)) \leqslant G(-\epsilon) + \left \| G(-2\epsilon) - G(-\epsilon) \right \|
\end{equation}
where both  $G(-\epsilon)$ and $G(-2\epsilon)$ are calculated by solving IVP defined in Eq. \eqref{26}. 

\subsection{Model Architecture and Implementation of STRODE}
\begin{figure}[!htb]

    \centering
    \includegraphics[width=0.55\textwidth]{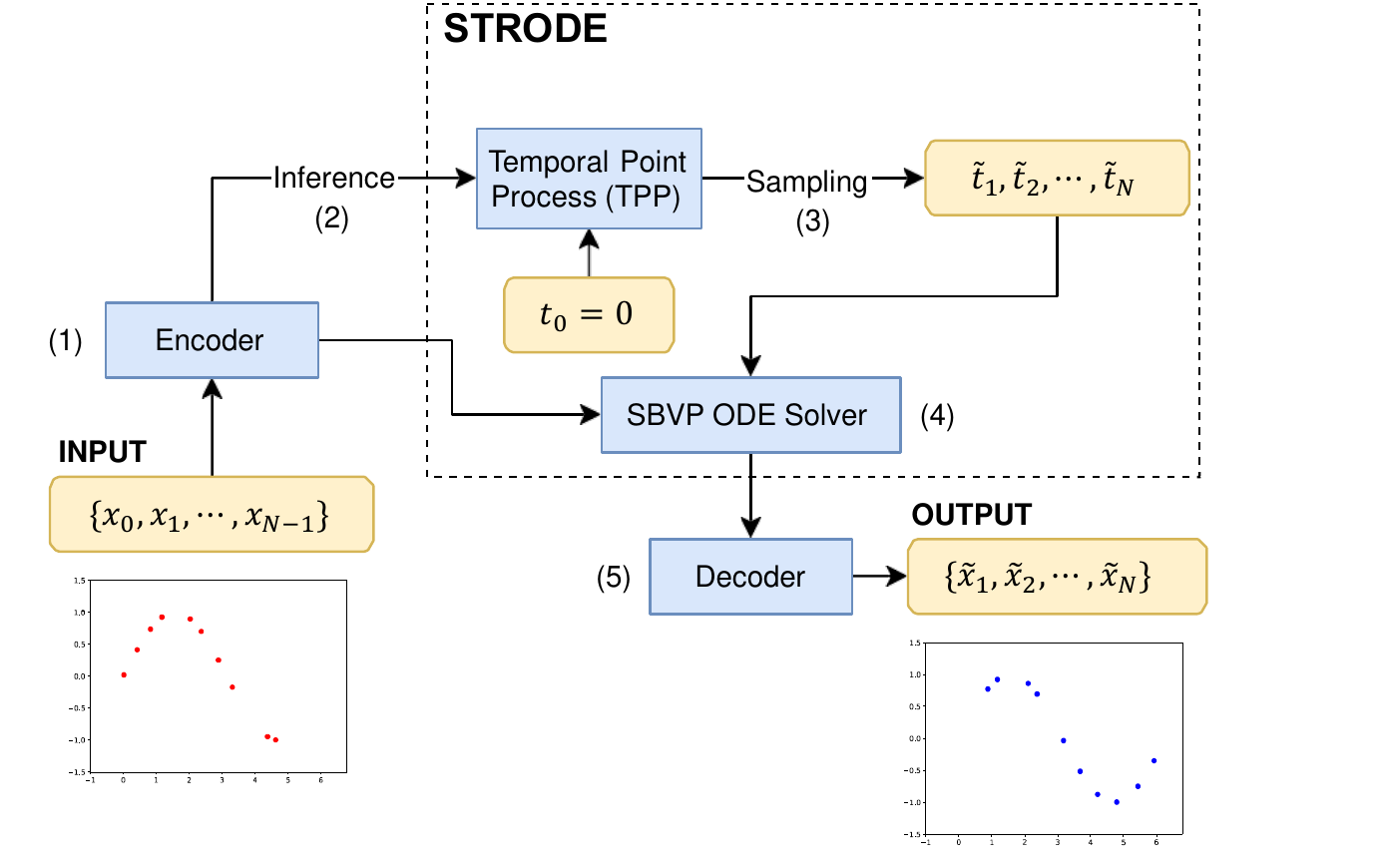}
    \caption{Architecture of STRODE for the toy dataset}
    \label{fig:exp1_figure}
    
\end{figure}

In the previous sections, we have discussed using STRODE to describe latent dynamics and infer boundary times of irregularly sampled time series data. Conceptually,  both the ODE and the boundary conditions of SBVP are defined over time series data itself. This can be an inappropriate way to model highly structured irregularly sampled high-dimensional data, such as video thumbnails or previews \cite{liu2015multi}, which is characterized by complex dependencies. 

STRODE can be conveniently extended to handle such data by introducing an encoder $f_{\theta_e}$ to describe its boundary conditions, which can be written as:
\begin{align}
  \{ h(0) =f_{\theta_e}({x}_{0}), h(\tilde{t}_{1})=f_{\theta_e}({x}_{1}),. . ., h(\tilde{t}_{N})=f_{\theta_e}({x}_{N}) \} 
\end{align}
such that the ODE in SBVP is capable of modeling complex dynamics of the hidden representations. 

In our experiment on the toy dataset (see Section \ref{exp:td} ), we use STRODE to handle irregularly sampled data from sine waves for an autoregressive task. Here, we provide details for  its implementation.  (see Supplement for STRODE implementations on other tasks).

Our STRODE consists of five major components: encoder, decoder, inference, sampling and SBVP ODE solver, which are shown in \figref{fig:exp1_figure}. The detailed implementation of such components are described as following:

\begin{enumerate}[(1)]
    \item \textbf{Encoder}:
    our encoder contains 2 fully connected layers and ReLU, each with 8 hidden nodes.
    \item \textbf{Inference}: the inference of the approximate posterior of the boundary time $q_i(t|x_{i})$ requires calculating the derivative of   ${\Phi_i}(t)$ with respect to $t$. In this experiment, ${\Phi_i}(t)$ is implemented as the neural network $f_{\theta_\Phi}$ (Eq. \eqref{sample22} ), whose architecture contains 2 fully connected layers, each with 16 hidden nodes and Tanh, whose outputs are further transformed into a scalar by another fully connected layer with Softplus. Then ${\Phi_i}'(t)$ is obtained by computing the derivative of  the neural network $f_{\theta_\Phi}$ with respect to its input of $t$, using automatic differentiation \cite{paszke2017automatic}. To ensure that the approximate posterior $q_i(t|x_{i})$ is positive-valued, the neural network weights are constrained to be negative. In doing so, if a weight is updated to be a positive value during training,
    we replace it with zero.  Similarly, the corresponding prior $p_{i}(t)$ requires  $\phi_i(t)$, which is implemented by the neural network $f_{\theta_\phi}$. The architecture of such a neural network is similar to that of $f_{\theta_\Phi}$, except that its neural network weights are constrained to be positive during training. To do this, if a weight is updated to be a negative value during training, we replace it with zero.

    \item \textbf{Sampling}: the boundary time samples are sequentially generated through  adopting Eq. \ref{sample22}.
    \item \textbf{SBVP ODE Solver}: with boundary time samples, we follow Eq. \ref{IVP_so} to obtain the prediction at $\tilde{t}_{i}$:
    \begin{equation}
        \tilde{x}_i = \tilde{h}(\tilde{t}_{i}) = \mathrm{ODESolve}(f_{\theta_o},h(\tilde{t}_{i-1}), \tilde{t}_{i-1},\tilde{t}_{i} )
    \end{equation}
    where the neural network $f_{\theta_o}$  includes 2 fully connected layers, each with 8 hidden nodes and Tanh.
    \item \textbf{Decoder}: the implementation is similar to what we adopt for the encoder. The difference is that the last layer of the decoder transforms data from a high-dimensional space to a one-dimensional one.
\end{enumerate}

\subsection{Latency-free Postdictive Modeling with Evenly Sampled Data}

Neuroscience investigations suggest that our visual and auditory system can process information retroactively, such that the incoming sensory inputs can affect the earlier perception \cite{stiles2018you}. There are advantages to this process: the accuracy of ``prediction'' is reassured with sufficient future information to be integrated. For example,  in human speech processing, understanding a word aids in distinguishing its constituent phonemes from another \cite{gwilliams2018spoken}.  However, such a process is difficult to be incorporated into existing algorithms for acoustic modeling, as the temporal range of subsequent context is mostly unannotated. Furthermore,  such processes could results in higher latency than other acoustic models.  

Here we propose another STRODE variant, dubbed regenerative STRODE, where both the dynamics of ODE and the posterior point process are capable of restarting themselves, allowing the latent temporal range to be captured for every feature frame.  

To be specific, suppose we are given a sequence of acoustic features  $ \mathbf{X}= \{\mathbf{x}_{0}, . . ., \mathbf{x}_{N}\}$ and training labels $ \mathbf{Y}= \{\mathbf{y}_{0}, . . ., \mathbf{y}_{N}\}$. Given that our posterior point process  probabilistically restarts when each feature frame occurs, Eq.\eqref{sample22} for generating  the i-$th$ boundary time sample is rewritten as: 
 \begin{equation} 
   \tilde{t}_{i} =  {\Phi_i}(\tilde{t}_{0}+\epsilon_0)= f_{\theta_\Phi} (\tilde{t}_{0}+\epsilon_0, x_i) \end{equation}
   where $\epsilon_0$ is a small positive value. We set it as $1\times 10^{-6}$.
Then the corresponding postdictive temporal ranges for i-$th$ frame is  $[i, i+\tilde{t}_{i}]$. Let $t \in [i, i+\tilde{t}_{i}] $, we assume the hidden state $h(t)$ satisfies the following ODE in which the dynamics restarts at each feature frame:
\begin{equation}
    {h}'(t) = f_{\theta_o}(h(t-i),t-i )   ,\    h(i)= f_{\theta_e}(\mathbf{x}_{i})
\end{equation}
Then the ODE solution at time $i+\tilde{t}_{i}$ can be written as:   
    \begin{equation}
         \tilde{h}(i+\tilde{t}_{i}) = \mathrm{ODESolve}(f_{\theta_o},h(i+\tilde{t}_{i}), i,i+\tilde{t}_{i} )
    \end{equation}
Next, we use the ODE solution $\tilde{h}(i+\tilde{t}_{i})$  as an additional input of the acoustic model. Such design  allows our model to look-ahead without compromising  input latency. 


We further impose a prior distribution for the ODE solution $\tilde{h}(i+\tilde{t}_i)$, which can be treated as a way of injecting prior knowledge for solutions of ODE. For instance, we can use the Gaussian distribution $\mathcal{N}({\mu}_{0},{\sigma}_{0}^{2})$, whose mean ${\mu}_{0}$ and variance ${\sigma}_{0}^{2}$ are the output of a non-linear function $f_{\theta_p}$ of $x_{i+1}$, to constrain the future ODE solution:
\begin{equation}
     \tilde{h}(i+\tilde{t}_i) \sim \mathcal{N}({\mu}_{0},{\sigma}_{0}^{2}),
     \ where \ [{\mu}_{0},{\sigma}_{0}^{2} ] = f_{\theta_p}(\mathbf{x}_{i+1} )  
\end{equation}

For optimization, we simply add the log-likelihood of all ODE solutions $p(\tilde{h}(i+\tilde{t}_i))$ to the original ELBO:   
\begin{equation} \label{elbo2}
\begin{split}
   \sum_{i=0}^{N} \{  \mathbb{E}_{ t_i \sim q_i(t|\mathbf{x}_{i}) }   \log p( \mathbf{y}_i|\mathbf{x}_i, t_i)  \\ -\mathrm{KL} (q_i(t|\mathbf{x}_{i}) \| p_i(t)) + \log p(\tilde{h}(i+\tilde{t}_i))   \}
 \end{split}
\end{equation}

\section{Experiments}
\label{exp}
We evaluate our STRODE on both synthetic and real-world time-series data. More specifically, we conduct the preliminary experiments on irregularly sampled 1D synthetic time-series data and  synthetic video thumbnail data based on MNIST to demonstrate its effectiveness in learning both the latent dynamics and timings of the time-series data. We further evaluate our model on CHiME-5, a realistic conversational speech recognition dataset to explore STRODE's potential in posdictive modeling.


 \subsection{Toy Dataset } \label{exp:td}

    \begin{figure*}
     
    \begin{center}
        \resizebox{0.66\textwidth}{!}{\includegraphics{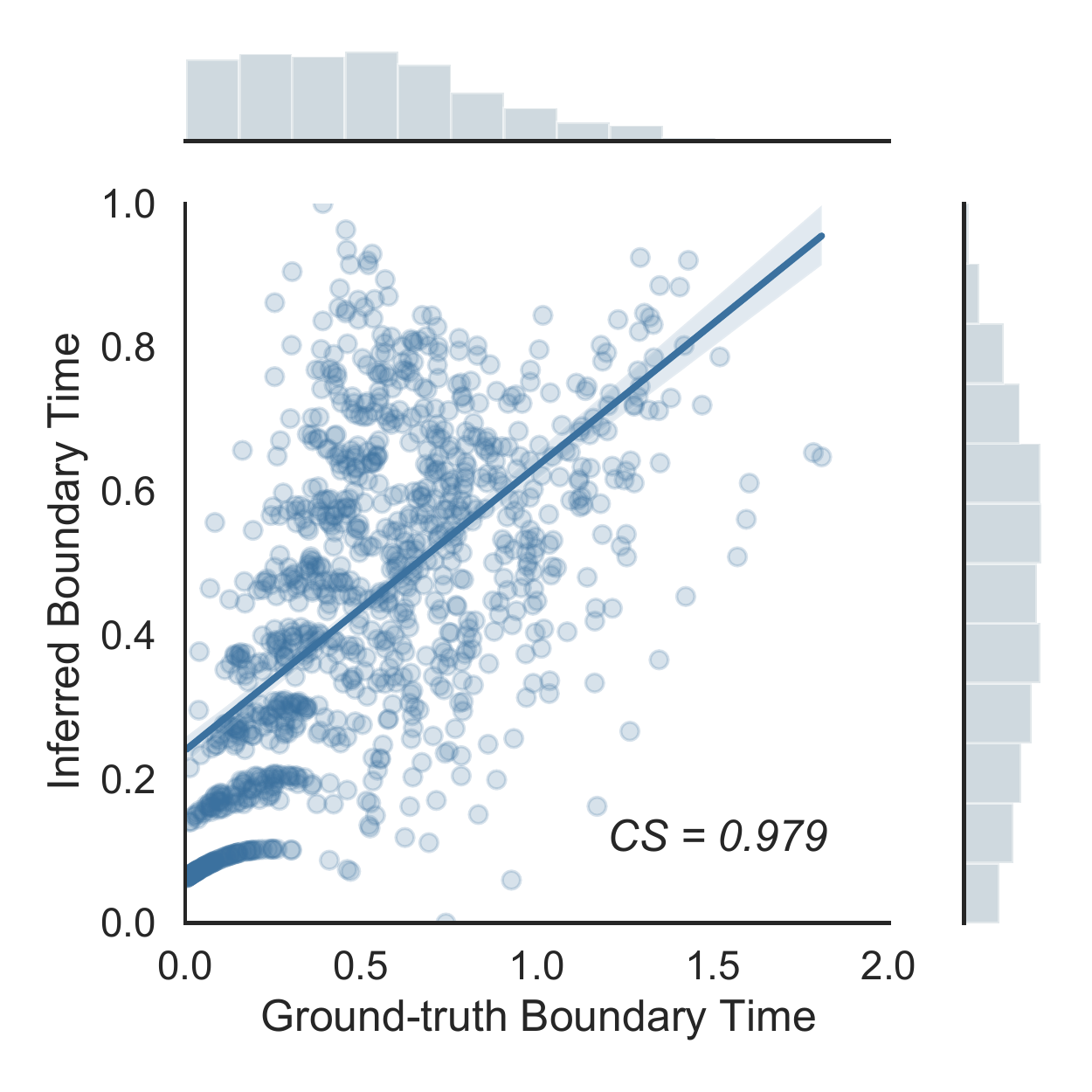}\includegraphics{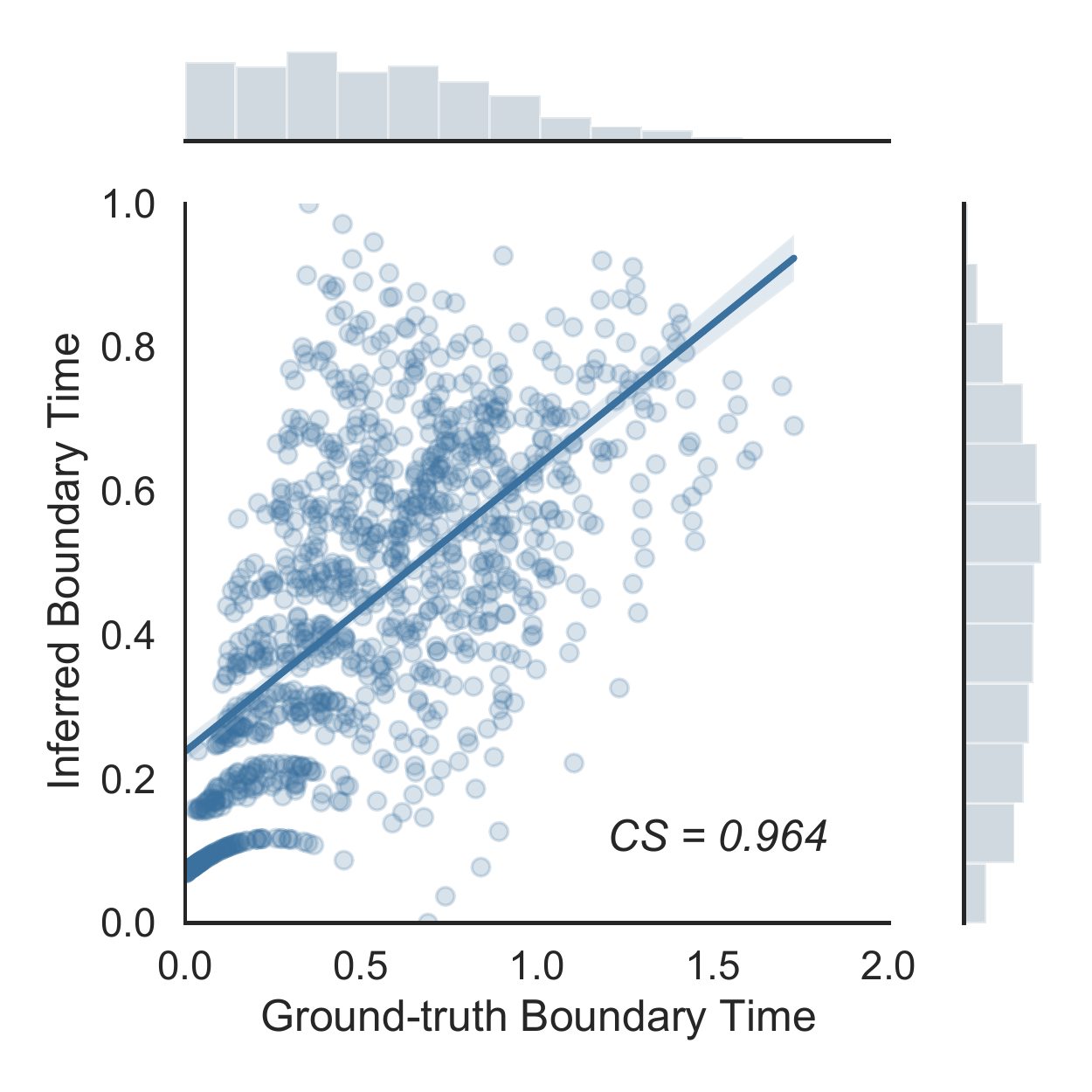}}
        \caption{Left: boundary time samples produced by STRODE for sine waves irregularly sampled with Hawkes process data;  Right: boundary time samples produced by STRODE for sine waves irregularly sampled with Poisson process data. }
        \label{fig:numerical}
    \end{center}
  
\end{figure*}

We start by conducting an autoregressive task using STRODE on a toy dataset with 1D sequential data.
We aim to demonstrate the capability of our STRODE in inferring informative boundary times given the time series data alone. Our datasets are generated by the sine function $\mathrm{A}\mathrm{sin}(wt)+\eta$, where the amplitude $\mathrm{A}=1$; the frequency $w=1.0$; $\eta$ represents Gaussian random noises; $t$ is the time point drawn from point processes. In this experiment, we consider two general point processes, namely a Poisson process with the conditional intensity $\lambda=10.0$ and a Hawkes process with the conditional intensity $\lambda=10.0$, $\alpha=0.5$, $\beta=1.0$. For each process, we create a dataset containing 5200 sequences with 10 different time points in each sequence. We use 5000 sequences for training 100 sequences for validation and 100 sequences for testing.  Training procedure of STRODE for this task are provided in the Supplement.

\subsubsection{Results and Analysis} 
We evaluate our model by calculating the cosine similarity (CS) between the inferred boundary time and the ground truth. We display the boundary time samples produced by STRODE for both Poisson process data and Hawkes process data shown in \figref{fig:numerical}. Note that we use min-max normalization to scale the boundary time samples between 0 and 1. We can see that STRODE achieves CS of 0.979 and 0.964 for the Hawkes process data and the Poisson process data respectively. This suggests that our model is capable of inferring timings of the irregularly sampled sine waves.
\subsection{Rotating MNIST Thumbnail: Evaluation on Irregularly Sampled High-dimensional Data without Timings }

We further study the generalization of STRODE for  irregularly sampled high-dimensional data. To do this, we construct  a synthetic video thumbnail dataset based on MNIST, the Rotating MNIST Thumbnail. 
We first generate 12000 video sequences of the rotating MNIST handwritten digits with constant angular velocity. We then adopt a point process and an exponential function to generate time points for selecting frames from video sequences. We generate the corresponding video thumbnails for each video sequence using a Hawkes process  with $\lambda=1.0$, $\alpha=0.5$, and $\beta=1.0$ and an exponential function $t=e^{a}+\epsilon$ where $a$ is evenly distributed and $\epsilon$ is normally distributed, resulting in two subsets of Rotating MNIST Thumbnail: Hawkes and Exponential.   We designate video thumbnails that are rotating by $0^{\circ}$ to  $180^{\circ}$ as the training and validation data, and assign video thumbnails rotating by $180^{\circ}$ to  $360^{\circ}$ as the test data.  For each subset, we generate 5000 video thumbnails for training, 1000 video thumbnails for validation and 1000 video thumbnails for testing.

 
\begin{table*}[htb!]
	\caption{ Cosine similarity (CS) (mean$\pm$ std) and MSE results on two subsets of Rotating MNIST Thumbnail}
	\label{tbl-rotateMNIST}

	\centering
	\begin{tabular}{l|cc|cc}
		\toprule
		DATASET & \multicolumn{2}{c|}{Hawkes} & \multicolumn{2}{c}{Exponential} \\
		& CS  & MSE ($\times10^{-3}$) & CS & MSE ($\times10^{-3}$) \\ 
		\midrule
		NODE \cite{chen2018neural}  & 0.907  & 6.66$\pm$0.03  & 0.923 & 7.69$\pm$0.02 \\
		ODE-RNN \cite{rubanova2019latent}  & 0.907   & 6.82$\pm$0.01 & 0.923 & 6.07$\pm$0.10 \\ 
		STRODE (Ours)  & 0.966$\pm$0.007  & \textbf{6.01}$\pm$\textbf{0.11} & 0.973$\pm$0.003 & 7.26$\pm$0.27 
		\\ 
		STRODE-RNN (Ours)  & \textbf{0.967}$\pm$\textbf{0.012}  & 6.35$\pm$0.14 & \textbf{0.974}$\pm$\textbf{0.005} & \textbf{5.94}$\pm$\textbf{0.03} \\ 
		\bottomrule
	\end{tabular}

\end{table*}

 \subsubsection{Models}
We follow implementation in \cite{rubanova2019latent} to incorporate STRODE to RNN. We then obtain a new variant of STRODE, dubbed as \emph{STRODE-RNN}.
We compare STRODE and STRODE-RNN with two baselines including: 
(i) NODE \cite{chen2018neural}, which in this task is an extension of neural ODE through adopting convolution layers in the encoder and deconvolution layers in the decoder.  
(ii) ODE-RNN \cite{rubanova2019latent}, which is an extension of ODE-RNN through adopting convolution layers in the encoder and deconvolution layers in the decoder as well.   
All baselines adopt the same architecture of encoder and decoder and take the same input as our STRODEs while having the similar number of parameters. 

\subsubsection{Training Procedure}

Similar to experiments on toy dataset, our STRODEs on Rotating MNIST Thumbnail is  trained by the ELBO, in which  the likelihood term is simplified by an MSE term, using the Adam optimizer with a learning rate in the range $[2\times10^{-4}, 6\times10^{-4}]$. 
We follow the training strategies of $\beta$-VAE \cite{higgins2017beta} to reweight the importance of  the $\mathrm{KL}$ term. The range of the $\mathrm{KL}$ term is from  $1 \times 10^{-5}$ to $1 \times 10^{-4}$. 
We repeat this training procedure across 3 different random
seeds.


 \subsubsection{Results and Analysis}
We report test-set CS and MSE on both Hawkes and Exponential subsets of  Rotating MNIST Thumbnail, as shown in \tabref{tbl-rotateMNIST} \footnote{We find that results differ when using different GPUs. We, therefore, rerun the experiments with a NVIDIA TESLA V100 GPU and update the results in \tabref{tbl-rotateMNIST}.}. As timings of each video frame are unknown, both NODE and ODE-RNN set the difference between the initial time and terminate time of the ODE solvers as a constant (i.e.$=1$). As such, a sequence of natural numbers are adopted when calculating CS with ground truth boundary time for both NODE and ODE-RNN. For the Hawkes task, we can see that STRODE and STRODE-RNN outperforms other baselines in terms of CS score; our STRODE archives the lowest MSE. This suggests that our STRODEs are capable of inferring both timings and complex dynamics of high-dimensional time-series data. 
We further apply our STRODEs to handle the video thumbnails drawn from exponential functions. 
Again, our STRODEs perform much better than baselines in terms of CS score, which demonstrates the expressive power of STRODEs in handling data whose generation doesn't follow a TPP. 


\subsection{CHiME-5: Latency-free Postdictive Modeling with Evenly Sampled Data }
 
CHiME-5 was originally designed for the CHiME 2018 challenge \cite{barker2018fifth}. It contains a very challenging problem of conversational speech recognition in everyday home environments. For a fair comparison with \cite{huang2020deep}, only the audio data recorded by binaural microphones is employed for training and evaluation in this experiment. The Train, Dev and Eval include about 40 hours, 4 hours, and 5 hours of real conversational speech respectively. The evaluation was performed with a trigram language model trained from the transcription of CHiME-5.

\subsubsection{Models}
We adopted SRU \cite{lei2017simple} as the building block to construct all RNNs involved in both our STRODE and baselines. 
Note that Latent ODE \cite{rubanova2019latent} is not adopted as one of our baselines, as it employs a seq2seq architecture, which is not applicable for HMM-based acoustic modeling. For simplicity, we denote regenerative STRODE as STRODE in this task. We compare our STRODE with the following baseline models:
(i) ODE-RNN with 9 stacked SRU layers in the encoder and 9 stacked SRU layers in the decoder  (ii) RTN  with a 6-layer SRU encoder and a 9-layer SRU decoder. 
 
All baselines except RTN take the same input as our STRODE. RTN takes multiple utterances as input, which contains significantly more contextual information than our STRODE and other baselines. To ensure similar numbers of model parameters for different models, we set the number of RNN hidden states per layer to 1,100 for  ODE-RNN and 900 for STRODE.  The size of ODE solutions for both ODE-RNN and STRODE 
is set as 128. More implementation details of STRODE are provided in the Supplement.

\subsubsection{Training Procedure}
For a fair comparison, we adopted the same configuration with \cite{huang2020deep} to train all GMM-HMM.  They were then used to derive the state targets for subsequent DNN acoustic model training through forced alignment for Train of CHiME-5. The state targets of CHiME-5 for subsequent STRODE-based and baseline acoustic model training were obtained by aligning the training data with the DNN acoustic model through the iterative procedure outlined in \cite{iterative_procedure}. Our STRODE and baselines on CHiME-5 are  trained by maximizing the ELBO, using BPTT \cite{werbos1990backpropagation} and SGD with learning rates ranging from $0.13$ to $0.19$. We apply a dropout rate of 0.1 to the connections between neural network layers except that of ODE solvers. We reweight the importance of both $\mathrm{KL}$ terms and the prior log-likelihood. Both weights are set as $1\times10^{-3}$. 
\begin{table}[htp!]
  \caption{Model configurations and the training time for CHiME-5. N: number of hidden states per layer;   P: number of model parameters; T: Training time per epoch (hr).}
\label{tbl-chime5_time}
\centering
\begin{footnotesize}
  \begin{tabular}{lccc}
    \toprule
{\sf Model}   & {\sf N} & {\sf P} & {\sf T} \\ 
    \midrule


ODE-RNN \cite{rubanova2019latent}  & 1100 & 77M & 0.6 \\
RTN \cite{huang2020deep}                                                &1024            & 70M                                & 0.3                     \\
STRODE (Ours) & 900 & 76M & 0.7 \\
    \bottomrule
  \end{tabular}
  \end{footnotesize}
  \vskip -0.08in
\end{table}

\subsubsection{Results and Analysis}

\paragraph{Training Speed and Model Complexity of STRODE}
Table~\ref{tbl-chime5_time} shows the configurations of baseline models and the new STRODE model for CHiME-5. The training time per epoch for CHiME-5 is also reported.
In our experiments, the timing experiments use PyTorch package and are performed on Ubuntu 16.04 with a single Intel Xeon Silver 4214 CPU and a GTX 2080Ti GPU. Each model takes around 20 iterations, with their average running time reported. We can see that STRODE runs as fast as ODE-RNN with a similar number of parameters. The running time of STRODE is comparable to RTN, though it runs 2.5 times faster than ours.  
\begin{figure*}[t!]
\vskip -0.1in
    \begin{center}
        \resizebox{1.0\textwidth}{!}{\includegraphics{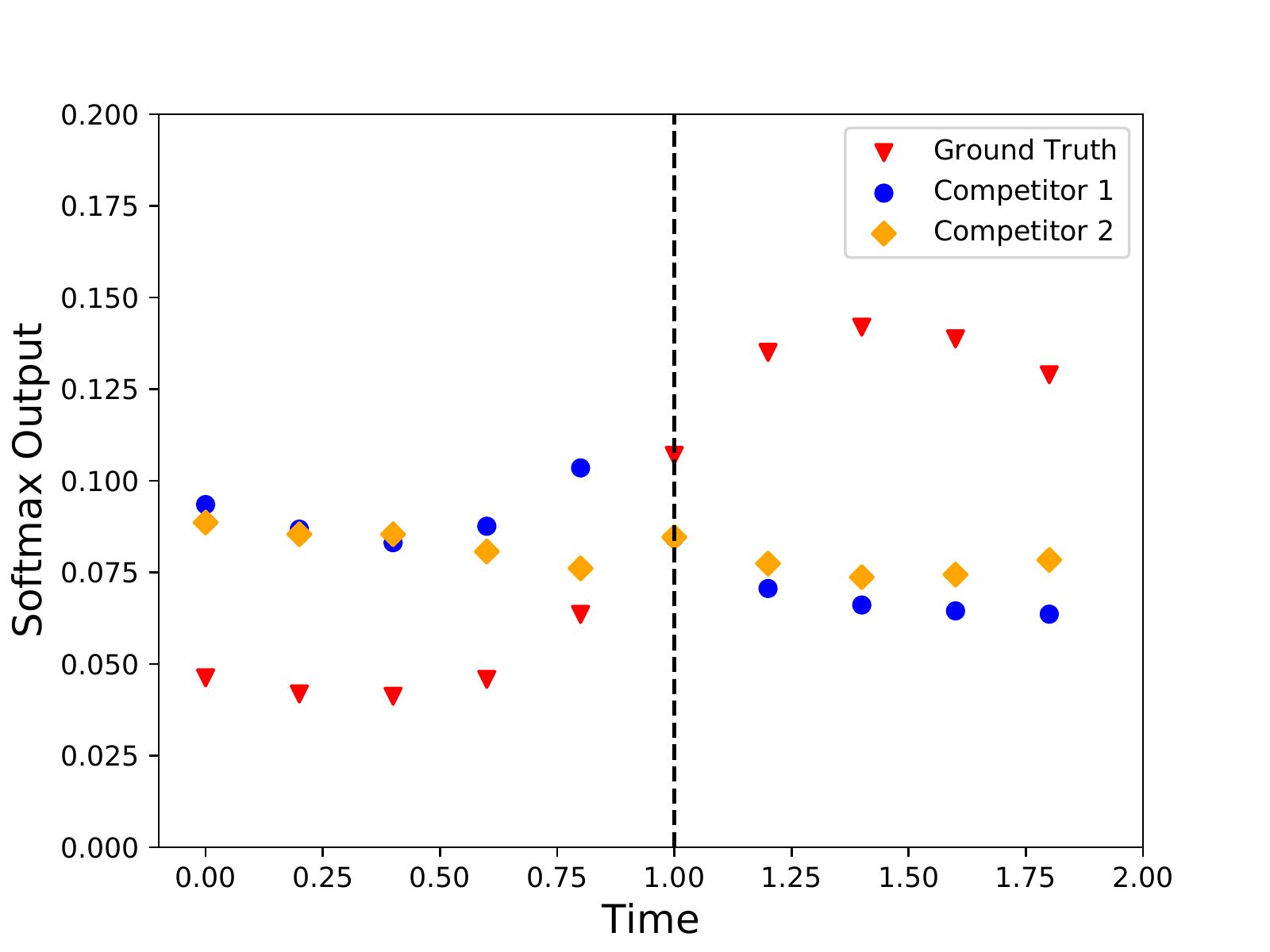}
        \hskip -1.2cm
        \includegraphics{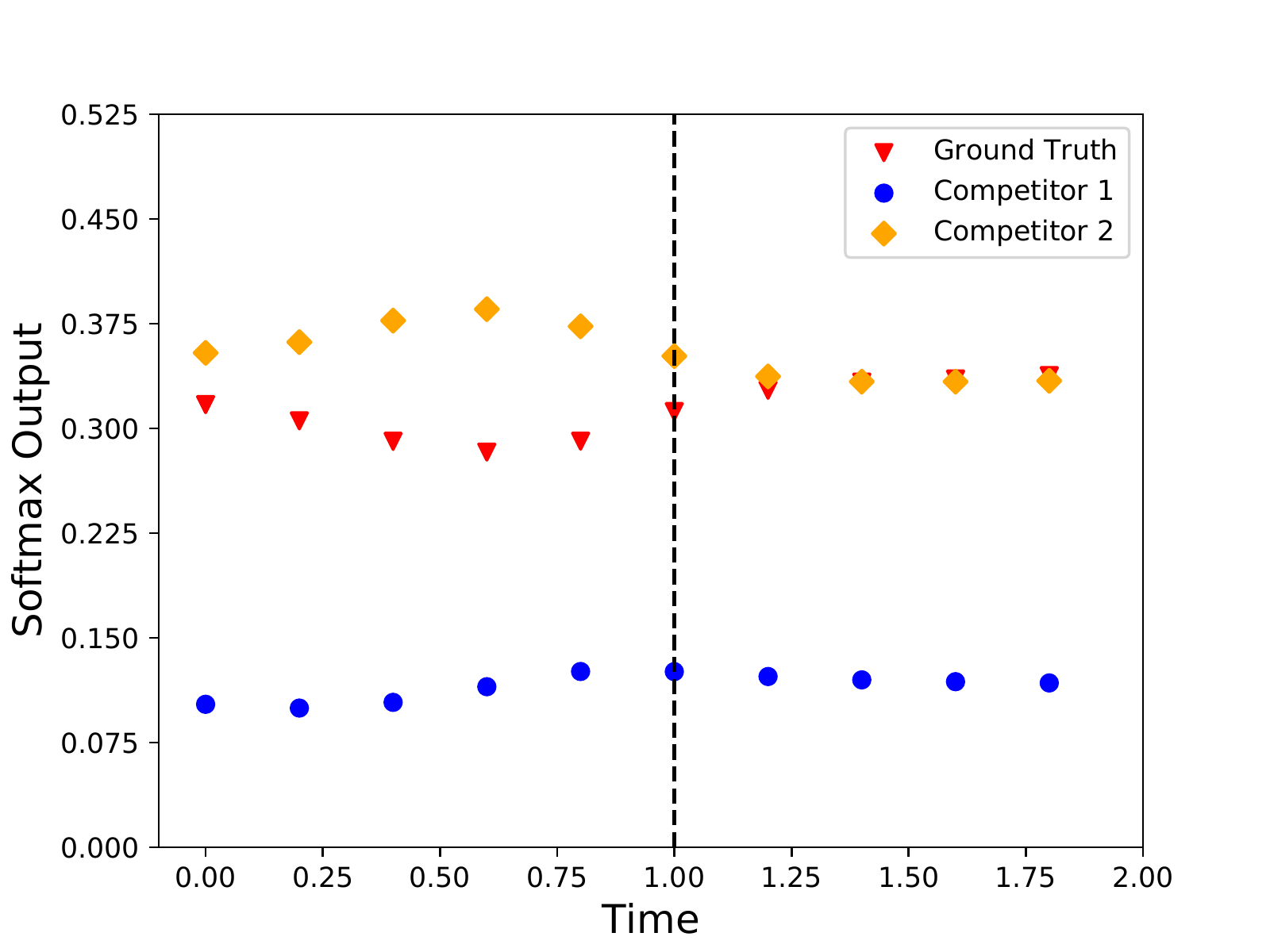}
        \hskip -1.2cm
        \includegraphics{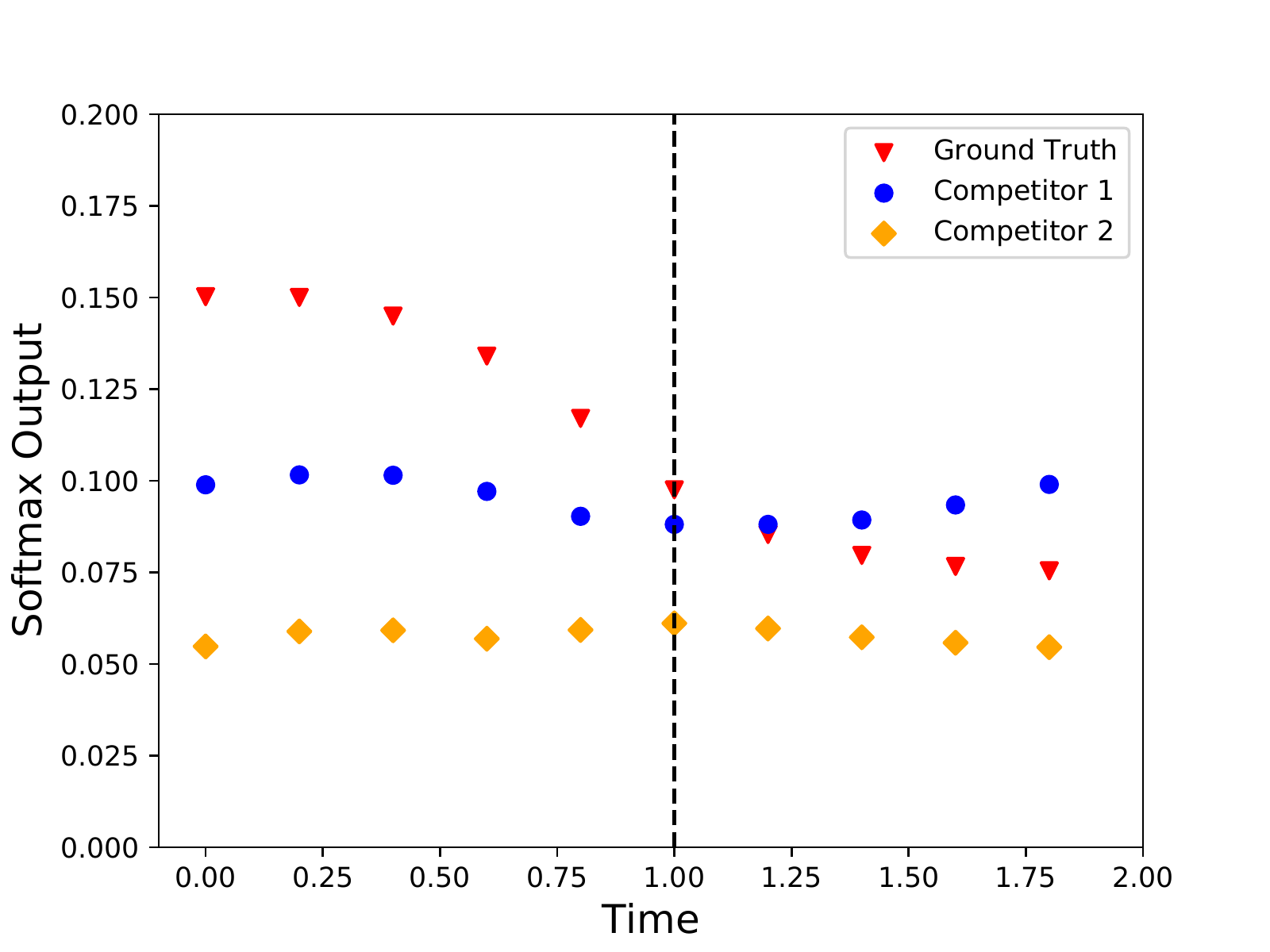}}
        \vskip -0.2cm
        \caption{The Softmax outputs  by taking the ODE solutions $\{ h(i+t)\}$ at future time points as an extra input of the acoustic model. The dotted line  corresponds to the original Softmax output of STRODE.  Both Fig. 3(left) and Fig. 3(middle) provide examples that match the assumption of postdictive modeling, while Fig. 3(right) provides some counter examples.   }
        \label{fig:ASR}
    \end{center}
\vskip -0.1in    
\end{figure*}

\paragraph{Quantitative evaluation of STRODE on CHiME-5}
Table~\ref{2021-tbl-chime5} shows the word recognition performance of the baseline models and the new STRODE model for CHiME-5.  It is observed that ODE-RNN outperforms the DNN baseline from Kaldi s5 \cite{kaldi}. STRODE performs the best among all models in terms of WER, outperforming the ODE-RNN by 2.7\%. Compared with the state-of-the-art acoustic
RTN model, our STRODE achieves 0.9\% absolute WER reduction even though the input of our STRODE contains much less contextual information than RTN.
\begin{table}[htp!]
   
	\caption{WER (\%) on eval of CHiME-5. }
	\label{2021-tbl-chime5}
	\centering
	\begin{tabular}{lll}
		\toprule
		
		{\sf Model}   &WER  \\ 
		\midrule
		Kaldi DNN \cite{kaldi}             & 64.5            \\ 
		ODE-RNN \cite{rubanova2019latent} & 59.0 \\
		RTN \cite{huang2020deep}               & 57.4       \\ 
		STRODE (Ours)     & \textbf{56.3} \\
		\bottomrule
	\end{tabular}
\vskip -0.2in

\end{table}

\paragraph{Qualitative evaluation of Postdictive Temporal Range Produced by STRODE}
We generate postdictive temporal range $[i, i+\tilde{t}_{i}]$ for randomly selected feature frames from utterances in Dev of CHiME-5. For $t\in [0, 2\tilde{t}_{i}]$,  we obtain a sequence of ODE solutions $\{ h(i+t)\}$ at future time points, and produce the Softmax output  by taking $\{ h(i+t)\}$ as an extra input of the acoustic model. In \figref{fig:ASR}, we only display the Softmax outputs for the ground truth label (red), and the others are from the class labels with the top 2 highest Softmax output values.  
It is evident from \figref{fig:ASR}(left) that 
the Softmax outputs for ground truth label increases over time, and tops exactly after $i+\tilde{t}_{i}$. Such a pattern matches the assumption of postdictive modeling. 
For the \figref{fig:ASR}(middle), STRODE fails to produce accurate results without integrating sufficient information. Interestingly, the STRODE hits the target after $i+\tilde{t}_{i}$, which suggests that our model requires a longer context for targeting the ground truth label. We also show some counter examples \figref{fig:ASR}(right) in which STRODE fails to produce accurate results when more contextual information is integrated.  

\section{Conclusion}
We propose a novel neural ODE model named STochastic boundaRy ODE (STRODE) for handling time series data without timing annotation during training.  We provide theoretical guarantees on the learning of STRODE. We show that our model is capable of inferring  timings and the dynamics of time series data without requiring any timing annotations during training. We demonstrate that our STRODE can be applied to address postdictive modeling. Our experiments on CHiME-5 show that our method outperforms ODE-RNN acoustic model in ASR.

\section*{Acknowledgments}
The authors would like to thank the anonymous reviewers for their
insightful comments and suggestions, Ling Qin and Kai Song for their assistance in proofreading the initial manuscript.
This project was partially funded by research grant R-252-000-B78-114
from the Ministry of Education, Singapore.


\nocite{langley00}
\bibliography{abbrev,mybib}
\bibliographystyle{icml2021}
\end{document}


\language0
\lefthyphenmin=2
\righthyphenmin=3

\maketitle

\section{Proof of Lemma 1}

\begin{lemma}\label{thm:t1}
Let $\epsilon$ be a positive real constant. Let $U\subset R^{n}$ be an open set. Let $f_1,f_2:[a-2\epsilon,a)\to R^{n}$ be a continuously differentiable function and $\left \| {f_1}' \right \| \leqslant M$ where $M$ is a positive constant.   Let $y_1,y_2:[a-\epsilon,a)\to U$ satisfy the initial value problem: 
\begin{align*}
{y}'_{1} &= f_1(t), \ \ y_1(a-\epsilon)=x_1 \\
{y}'_{2} &= f_2(t) = f_1(t-\epsilon),\ \ y_2(a-\epsilon)=x_2 
\end{align*}
Suppose both $x_1$ and $x_2$ depend on $\epsilon$. As $\epsilon \rightarrow 0$, we have: 
\begin{equation*}
\lim_{\epsilon \rightarrow 0} 
\left (\lim_{t \rightarrow a} { \left \| {y}_{1}(t) - {y}_{2}(t) \right \|}\right ) \leqslant  \lim_{\epsilon \rightarrow 0}  \left \| x_1 - x_2 \right \|
\end{equation*}
\end{lemma}

\begin{proof}

For any constant $C \geqslant  0$, we have:
\begin{equation} \label{1}
 \left \| f_1(t) - f_1(t) \right \| \leqslant C \left \| x_1 - x_2 \right \|
\end{equation}
Let constant $K \geqslant  1$, we have:  
\begin{align} \label{2}
    \left \| f_1(t) - f_2(t) \right \| &= 
    \left \| f_1(t) - f_1(t-\epsilon) \right \| \nonumber \\ 
    &\leqslant K  \left \| f_1(t) - f_1(t-\epsilon) \right \|
\end{align}

Due to Eq.\eqref{1} and Eq.\eqref{2}, two assumptions of Gronwall’s Inequality (Theorem 2.1) \cite{howard1998gronwall} are met. Then let $C=0$, for any $t\in [a-\epsilon,a)$, we have:
\begin{align*} 
 \left \| {y}_{1}(t) - {y}_{2}(t) \right \| &\leqslant 
   \left \| x_1 - x_2 \right \| +   
\underbrace{ \int_{a-\epsilon}^{t} K  \left \| f_1(s) - f_1(s-\epsilon) \right \|  ds }_{(a)}
\end{align*} 
We then take part (a). There exists a $\theta_s \in (0, \epsilon) $ such that as $\epsilon \rightarrow 0$, we have: 
\begin{align} \label{3}
\lim_{\epsilon \rightarrow 0}\left ( 
\lim_{t \rightarrow a}{ \int_{a-\epsilon}^{t}   K  \left \| f_1(s) - f_1(s-\epsilon) \right \|  ds }\right ) &= 
\lim_{\epsilon \rightarrow 0}  { \int_{a-\epsilon}^{a} K  \left \| f_1(s) - f_1(s-\epsilon) \right \|  ds } \nonumber \\
&=  \lim_{\epsilon \rightarrow 0}  { \int_{a-\epsilon}^{a}  K \epsilon \left \|  {f_1}'(s-\theta_s) \right \|   ds}  \nonumber \\
&\leqslant \lim_{\epsilon \rightarrow 0}  {   K \epsilon^{2}M  } \nonumber =0
\end{align}

Then we have:
\begin{align}
\lim_{\epsilon \rightarrow 0} 
\left (\lim_{t \rightarrow a} { \left \| {y}_{1}(t) - {y}_{2}(t) \right \|}\right ) 
&\leqslant    \lim_{\epsilon \rightarrow 0}\left ( 
\lim_{t \rightarrow a} \left ( {  \left \| x_1 - x_2 \right \|  + \int_{a-\epsilon}^{t}   K  \left \| f_1(s) - f_1(s-\epsilon) \right \|  ds }\right )   \right )                    \\ 
&\leqslant \lim_{\epsilon \rightarrow 0}  \left \| x_1 - x_2 \right \|    
\end{align}

\end{proof}

\section{Proof of Theorem 1}

\begin{theorem}\label{thm:t2}
Suppose we are given two arbitrary distributions, $q(t)$ and $p(t)$ with $t \in [0,+\infty)$. Let $ m=-e^{-t}$. Let $\epsilon$ be a positive real constant and let $g:[-1,0)\to R $ be a continuous function. There exists a $G:[-1,0)\to [0, +\infty)$ that satisfies an initial value problem: 
\begin{align*}
 &{G}'(m) = g(m),\ \ G(-1)=0 
 \nonumber
 \\ &where \  g(m) = \frac{-q(- \log(-m)) }{m} \log  \frac{ q(- \log(-m))}{p(- \log(-m))}
\end{align*}
Such that as $\epsilon \rightarrow 0$,  we have:
\begin{equation}
\lim_{\epsilon \rightarrow 0} \left (\mathrm{KL}(q(t) || p(t)) \right ) \leqslant \lim_{\epsilon \rightarrow 0} \left (G(-\epsilon) +   \left \| G(-2\epsilon) - G(-\epsilon) \right \|\right ) 
\nonumber
\end{equation}

\end{theorem}

\begin{proof}
Let $m \in [-1,0)$ and let $t = -\log(-m) $, we have:
\begin{align*}
\mathrm{KL}(q(t) || p(t)) &= \int_{0}^{+\infty}q(t)\log(\frac{q(t)}{p(t)})dt \\
&=  \lim_{l\rightarrow 0} \int_{-1}^{l} \underbrace{ \frac{-q(- \log(-m)) }{m} \log  \frac{ q(- \log(-m))}{p(- \log(-m))} }_{g(m)} dm
\end{align*}

With the integrand $g(m)$, we then construct a $G:[-1,0)\to [0, +\infty)$ that satisfies an initial value problem: 
\begin{equation}
 {G}'(m) = g(m),\ \ G(-1)=0 
 \nonumber
\end{equation} 
Then we have:
\begin{equation}
\mathrm{KL}(q(t) || p(t)) = \lim_{l\rightarrow 0} G(l) = \lim_{l\rightarrow 0} \int_{-1}^{l} {g(m) dm} 
 \nonumber
\end{equation}
Since $g(m)$ is not analytic at the point 0, we separate the solution $\lim_{l\rightarrow0}G(l)$ into two parts:
\begin{align} \label{4}
  \mathrm{KL}(q(t) || p(t)) &=  \int_{-1}^{-\epsilon} {g(m) dm}  + \lim_{l\rightarrow 0} \int_{-\epsilon}^{l } {g(m) dm} \nonumber \\ &= G(-\epsilon)  + \lim_{l\rightarrow 0} \int_{-\epsilon}^{l } {g(m) dm}
\end{align}

Let $G_1,G_2:[-\epsilon,0)\to [0, +\infty)$ satisfy the initial value problem: 
\begin{align*}
{G}'_{1} &= g(m), \ \ G_1(-\epsilon)=G(-\epsilon) \\
{G}'_{2} &= g(m-\epsilon),\ \ G_2(-\epsilon)=G(-2\epsilon) 
\end{align*}
By Lemma \ref{thm:t1}, as $\epsilon \rightarrow 0$  we have: 
\begin{align} \label{5}
\lim_{\epsilon \rightarrow 0} \left (
{\lim_{l\rightarrow 0} \int_{-\epsilon}^{l } {g(m) dm}}\right )
&=
\lim_{\epsilon \rightarrow 0} \left (
\lim_{m \rightarrow 0} { \left \| {G}_{1}(m) - {G}_{2}(m) \right \|}\right ) \nonumber \\ 
& \leqslant   \lim_{\epsilon \rightarrow 0}
\left \| G(-\epsilon) - G(-2\epsilon) \right \|
\end{align} 

Combining Eq.\eqref{4} and Eq.\eqref{5}, we have:
\begin{align} 
  \lim_{\epsilon \rightarrow 0} \left (
  \mathrm{KL}(q(t) || p(t))\right )  &= 
  \lim_{\epsilon \rightarrow 0}  (
  G(-\epsilon)  + \lim_{l\rightarrow 0} \int_{-\epsilon}^{l } {g(m) dm} ) \\
  &\leqslant   \lim_{\epsilon \rightarrow 0} \left (
  G(-\epsilon)  +  \left \| G(-\epsilon) - G(-2\epsilon) \right \|\right )
\end{align}

\end{proof}

\section{Experiments}

In this section, we provide more details for experiments. Notably, any ODE involved in this experiment are solved by the Euler method with step size 0.1 (such hyperparameter can be tuned to further improve performance).

\subsection{Training Procedure of STRODEs on Toy Dataset}
Our STRODE on toy dataset is  trained by maximizing the ELBO, in which  the likelihood term is simplified by an MSE term. We adopt the Adam optimizer with a learning rate of $4\times 10^{-4}$. We apply a dropout rate of 0.1 to the connections between neural network layers except that of ODE solvers.

\subsection{Detailed Implementation of STRODE on Rotating MNIST Thumbnail }
\begin{figure}[!htb]
    \centering
    \includegraphics[width=0.75\textwidth]{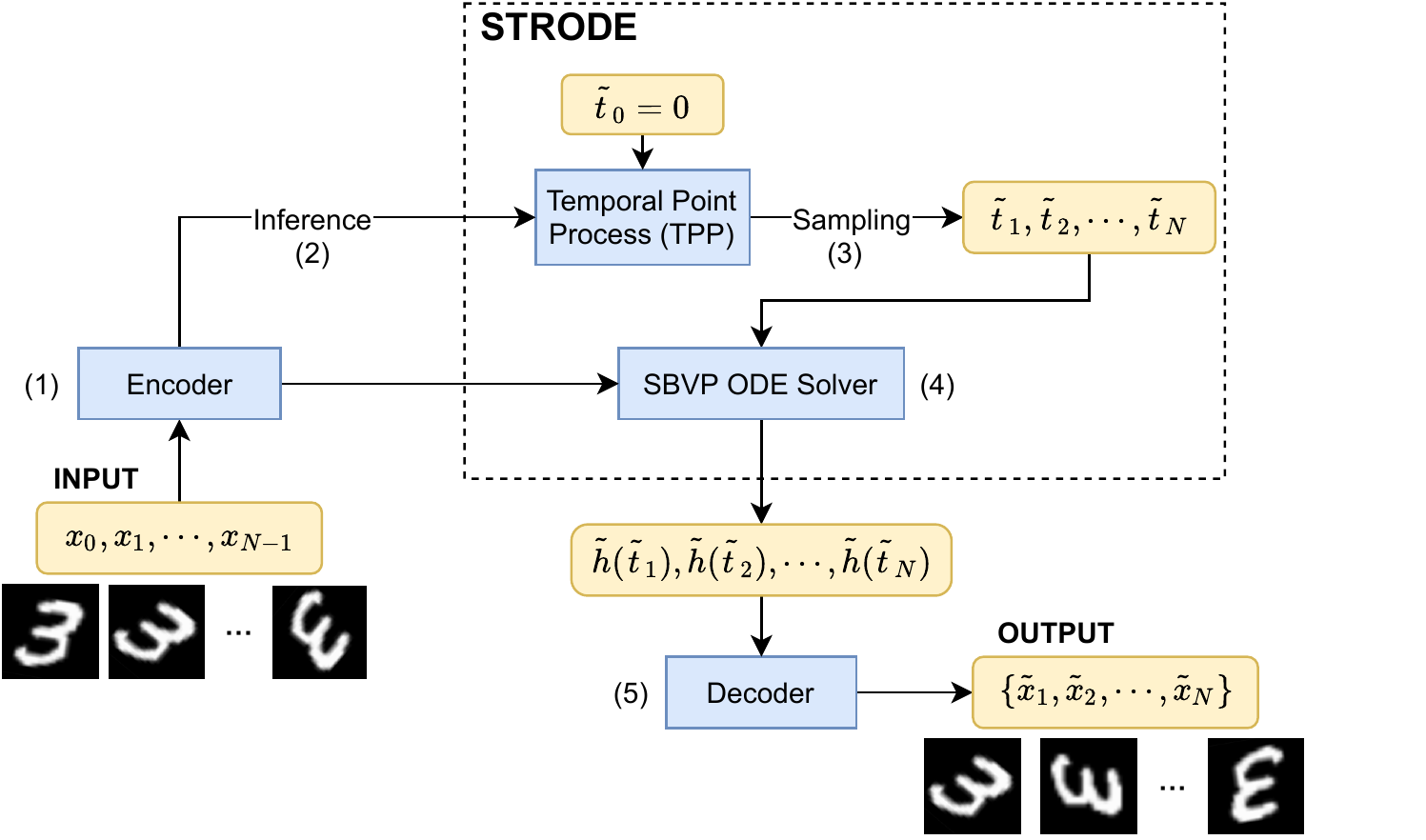}
    \caption{Architecture of STRODE for Rotating MNIST Thumbnail}
    \label{fig:exp2_figure}
\end{figure}

\begin{enumerate}[(1)]
    \item \textbf{Encoder}: our encoder for this task contains four convolution layers (kernel size: $5\times 5$) with the following input-output dimensions: $ 1\times64\times64  \rightarrow 128\times16\times16   \rightarrow 256\times8\times8   \rightarrow 512\times4\times4 \rightarrow 512\times1\times1 $, where numbers indicate \#feature maps$\times$width $\times$height. Batch-normalization is applied to each layer except the first and last ones. Activations are LeakyReLU.


    \item \textbf{Inference}: the implementation is similar to what we adopt on the toy dataset, except that the architecture of the neural network for both $\Phi_i(t)$ and $\phi_i(t)$ includes 2 fully connected layers, each with 128 hidden nodes and Tanh, whose outputs are further transformed into a scalar by another fully connected layer with Softplus.

    \item \textbf{Sampling}: the implementation is similar to what we adopt on the toy dataset.
    \item \textbf{SBVP ODE Solver}:
    the implementation is similar to what we adopt on the toy dataset.
    \item \textbf{Decoder}: 
     our decoder for this task contains four deconvolution layers (kernel size $5\times5$ ) with the following input-output dimensions: $ 512\times1\times1  \rightarrow 512\times4\times4   \rightarrow 256\times8\times8   \rightarrow 128\times16\times16 \rightarrow 1\times64\times64 $, where numbers indicate \#feature maps$\times$width $\times$height. We adopt a batch-normalization layer with ReLU activation  after each deconvolution layer except the last one. Tanh activation is applied after the last deconvolution layer.
\end{enumerate}




\subsection{Detailed Implementation of STRODE on CHiME-5}
\begin{figure}[!htb]
    \centering
    \includegraphics[width=0.75\textwidth]{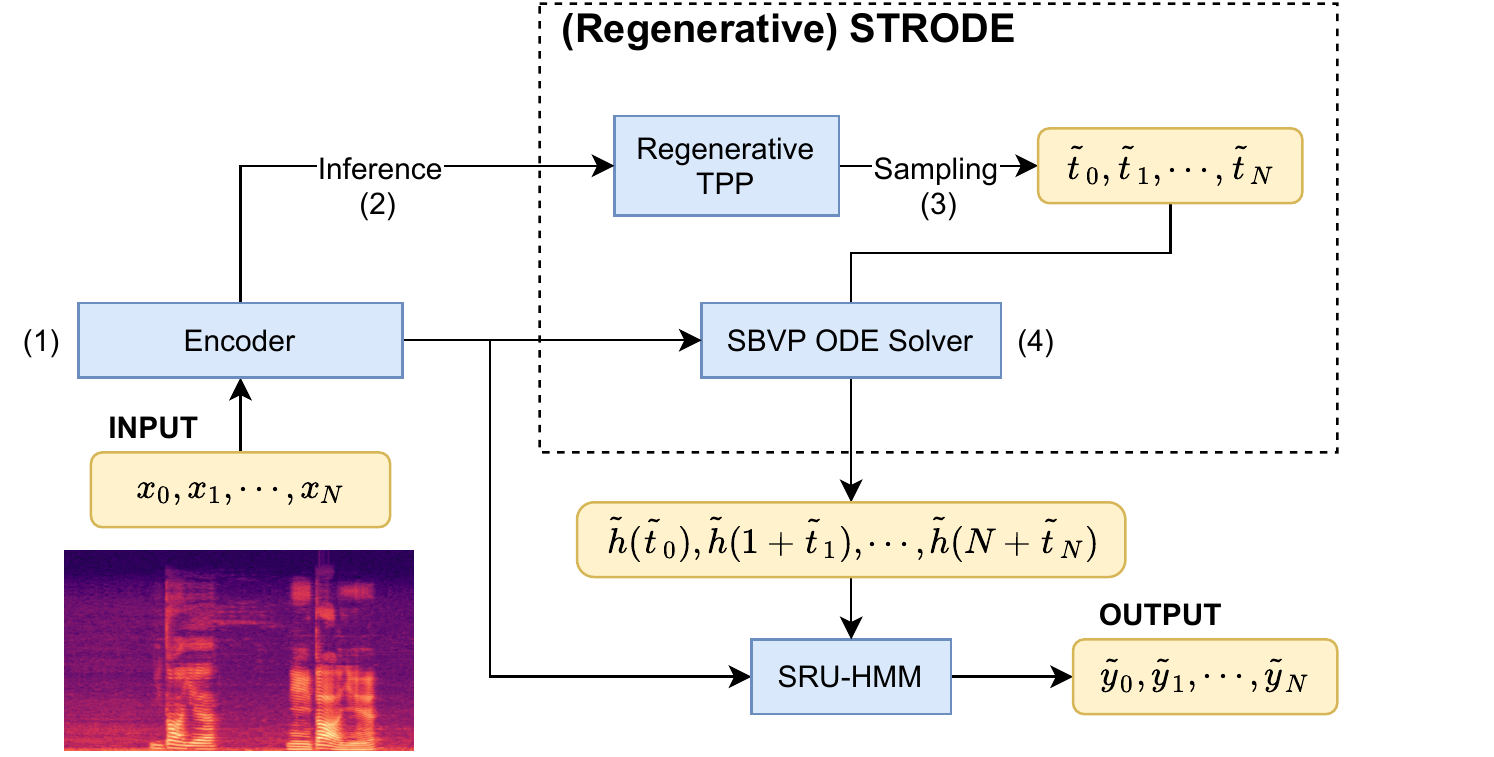}
    \caption{Architecture of STRODE for CHiME-5 }
    \label{fig:exp3_figure}
\end{figure}

\begin{enumerate}[(1)]
    \item \textbf{Encoder}: we adopt a 9-layer SRU to implement the encoder. Each SRU layer contains 900 hidden states. 
    
    \item \textbf{Inference}: the implementation is similar to what we adopt on the toy dataset, except that the architecture of the neural network for both $\Phi_i(t)$ and $\phi_i(t)$ includes 2 fully connected layers, each with 128 hidden nodes and Tanh, whose outputs are further transformed into a scalar by another fully connected layer with Softplus.

    \item \textbf{Sampling}:  the boundary time samples are sequentially generated through  adopting Eq. (34) in the main paper.

  \item \textbf{SBVP ODE Solver}:
    we follow Eq. (35) of the main paper to obtain the SBVP ODE solution for each frame, in which the neural network $f_{\theta_o}$  includes 2 fully connected layers, each with 64 hidden nodes and Tanh.

\end{enumerate}

\newpage

%



\clearpage
\bibliography{abbrev,mybib}
\bibliographystyle{apalike}

